# Medical Robots for Infectious Diseases:
# Lessons and Challenges from the COVID-19 Pandemic

Antonio Di Lallo, Robin R. Murphy, *Fellow, IEEE*, Axel Krieger, *Member, IEEE*, Junxi Zhu,
Russell H. Taylor, *Fellow, IEEE*, Hao Su\*, *Member, IEEE*

*Abstract*— Medical robots can play an important role in mitigating the spread of infectious diseases and delivering quality care to patients during the COVID-19 pandemic. Methods and procedures involving medical robots in the continuum of care, ranging from disease prevention, screening, diagnosis, treatment, and homecare have been extensively deployed and also present incredible opportunities for future development. This paper provides an overview of the current state-of-the-art, highlighting the enabling technologies and unmet needs for prospective technological advances within the next 5-10 years. We also identify key research and knowledge barriers that need to be addressed in developing effective and flexible solutions to ensure preparedness for rapid and scalable deployment to combat infectious diseases.

## I. Introduction

Since the first reports of a novel coronavirus (SARS-CoV-2) in December 2019, over 44.7 million patients have been infected worldwide, and more than 1.17 million patients worldwide have died from COVID-19, the disease caused by this virus (numbers as of October 29$^{th}$ 2020) [1]. Among them, 19% of infected persons were hospitalized, while 6% were admitted to the intensive care unit (ICU) [2]. Healthcare professionals acted as the frontline against the virus, resulting in a large exposure risk to infection and imperiling any mitigation efforts. The robotic community also took charge of an important role in providing aid to manage the pandemic [3, 4], and great efforts were made to adapt preexisting devices to the new challenges, which translated into a number of helpful solutions [5]. The shortage of time to design and develop ad-hoc robots pushed experts to reflect on the methods to get a ready response to future infectious disease crises, analyzing the challenges and opportunities for advancements in the technologies.

This paper covers the deployment of robots in the healthcare workflow across the continuum of care that goes from prevention, screening, and diagnosis to treatment and homecare, as depicted in Fig. 1. For each category of medical robots, the discussion starts from describing the current state of practice (i.e., robots deployed during the COVID-19 pandemic) and state of the art (i.e., research prototypes, not deployed during the COVID-19 pandemic), providing a review of the most advanced research progress. Then, we aim at providing a look ahead to a mid-term perspective, analyzing the major challenges and the enabling technologies that may be leveraged to make progress.

Reported systems are chosen according to the following inclusion criteria: 1) they are provided with some automated features that allow for at least a basic degree of autonomy, including shared autonomy or teleoperation; 2) they are either research prototypes or commercial products, have been already experimentally tested at least on a mockup (Technology Readiness Level greater than 3 according to the classification by the European Commission [6]); 3) they had or may have direct deployment in response to the COVID-19 pandemic.

With innovations in design, perception, actuation, and control, we envision that in the near future, robots may play a valuable role to assist the hospital personnel, relieve them from low-skilled or high-risk tasks, and to improve the quality of care of people who are ill or isolated because of infectious diseases.

## II. Clinical Background and Unmet Needs

COVID-19 is a respiratory viral disease with transmission via respiratory aerosols and micro-droplets. This places clinicians and healthcare professionals at risk of contracting the virus when caring for patients infected with COVID-19. The primary morbidity and mortality of COVID-19 are related to pulmonary involvement, and pneumonia is the primary cause of death in 44% of cases [2]. 15-20% of patients who develop COVID-19 will require ventilation in an intensive care unit (ICU) at some point during their illness [7]. Some of the most demanded resources during COVID-19 are healthcare workers, personal protective equipment (PPE), and ventilators. The infection risk for staff and the strain on PPE resources is exacerbated by the fact that healthcare workers must put on and take off PPE every time they enter an ICU or engage with a patient, even if only to perform a simple task such as changing a setting on a ventilator or changing the dosing of medication. Despite common traits, healthcare needs and responses were not identical everywhere. In many hard-hit countries, including the US, the COVID-19 pandemic has also ground to halt elective surgeries and routine health check-ups imperiling the public health and negatively impacting economic recovery [8].

A.D. Lallo, J. Zhu, H. Su are with Lab of Biomechatronics and Intelligent Robotics, Department of Mechanical Engineering, The City University of New York, City College, NY, 10031, USA.

R. Murphy is with Department of Computer Science and Engineering, Texas A&M University, College Station, TX, 77840, USA.

A. Krieger is with Department of Mechanical Engineering, The John Hopkins University, Baltimore, MD, 21218, USA.

R. Taylor is with Department of Computer Science, The Johns Hopkins University, Baltimore, MD, 21218, USA.

\*Corresponding author. Email: hao.su@ccny.cuny.edu

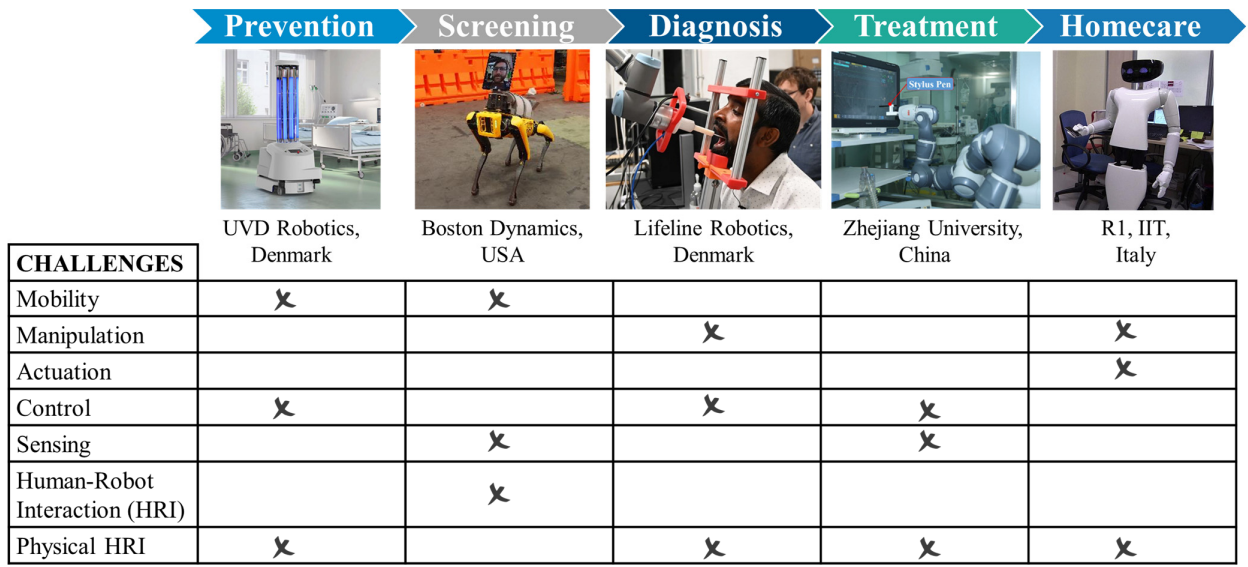

Fig. 1. In terms of continuum of care, five categories of representative medical robots were deployed during the COVID-19 pandemic to assist with healthcare needs associated to an infectious disease. For each category, the table outlines the three challenges (crossed cells) that according to the authors have the highest potential to enable key advances in that specific area. Further details are provided in the relative sections.

The COVID-19 pandemic is having a major impact on the global healthcare system, with telemedicine being one of the key drivers of the change. While the COVID-19 pandemic has driven the strong expansion of contact-less telemedicine use for urgent care and nonurgent care visits [9], further technological advances are necessary to expand the use of telemedicine to areas that require physical interactions such as for swab testing, imaging, nursing, or interventional treatment. Medical robots have the unique capability for bridging the gap between remote health care providers and patients by interacting with imaging and therapeutic equipment and with the patients, ushering in the next generation of telecare. But for robots to robustly and safely perform physical tasks such as swabbing a patient, changing a ventilator setting in an ICU, or performing an ultrasound scan, advancements in the areas of sensing, actuation, control, autonomy, and artificial intelligence (AI) are required.

### III. ROLE OF MEDICAL ROBOTICS IN INFECTIOUS ENVIRONMENTS

Robotics can play a key role in combating infectious diseases in four areas, including clinical care, logistics, reconnaissance, and continuity of work and maintenance of socioeconomic functions [3]. Here, we focus on the first area, within which we identify five categories of medical robots (Fig. 1) in the continuum of care, including prevention, screening, diagnosis (e.g., biological sampling and laboratory automation), treatment, and homecare (e.g., nursing).

#### A. Robots for Prevention

Disinfection is one of the key measures against infectious diseases. A common method adopted for the disinfection of public spaces, such as hospitals, is ultraviolet (UV) disinfection. It consists of exposing the surfaces to be disinfected to a specific ultraviolet light bandwidth, so-called UVC (200-280 nm), corresponding to the peak of the germicidal effectiveness [10]. In general, UV robots are comprised of a mobile base equipped with an array of lamps mounted on the top, spanning 360° coverage. The positioning of the device can be manual or autonomous. For instance, LightStrike Germ-Zapping Robot (Xenex, USA) is operated by trained hospital environmental services staff [11], while the UVD Robot (UVD Robots, Denmark, Fig. 1) relies on simultaneous localization and mapping to scan and navigate the environment on its own. As UVC light may be harmful to humans, the operation of these robots is typically suspended when opportune occupancy sensors detect the presence of a person in the space undergoing disinfection.

The possibilities offered by all these UV robots are strictly confined to the line of sight. To overcome this limitation, the University of Southern California developed a semi-autonomous mobile manipulator for ultraviolet disinfection by readapting its Agile Dexterous Autonomous Mobile Manipulation System (Fig. 2-a). It was endowed with UV-light wands, augmented vision guidance, and a teleoperation framework that relies on autonomous path planning algorithms to comply with high-level directives by a human operator. By exploiting human-in-the-loop control, it can handle targeted disinfection tasks through challenging scenarios that involve the approach to objects of interest and their manipulation.

Another class of robots uses chemicals to disinfect surfaces. Nanyang Technological University developed the eXtreme Disinfection roBOT (XDBOT, Fig. 2-b) that explores the environment and identifies objects to be disinfected using Light Detection and Ranging (LIDAR) and cameras. The wheeled mobile base supports a 6-axis robotic arm that handles an electrostatic sprayer and is remotely controlled by a human operator.

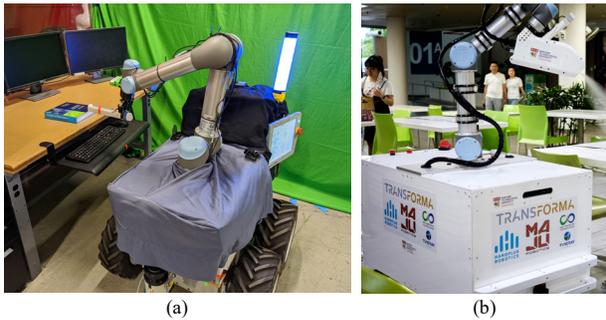

Fig. 2. Robots for disinfection. (a) ADAMMS (University of Southern California, USA) is a mobile robot with manipulation capabilities for UV disinfection. (b) XDBOT (Nanyang Technological University, Singapore) uses an electrostatic sprayer to carry out chemical disinfection.

As a common shortcoming, typically disinfection robots operate over a predefined temporal horizon while not providing any direct measurement of the decontamination evolution (e.g., mapping of the disinfection dose), which prevents the application of any feedback control on the accuracy of the process. Yet, factors like distance and orientation of surfaces play a crucial role in the effectiveness of the UV decontamination. Chemical decontamination with robots needs to consider or control multiple parameters, including concentration and quantity of disinfectant, contact time and temperature, residual activity and effects on material properties and surface roughness, and pH scale and interactions with other compounds. Future research is needed to understand how to control disinfection with robotic systems better, possibly exploiting feedback to enhance the reliability of the process. Other opportunities are related to the development of AI to turn the current semi-autonomous devices into fully autonomous robots with enhanced efficiency.

*B. Robots for Screening*

Early identification of infected persons is a critical function in the management of infectious diseases. In hospitals, triage is the first step to receive people who need medical attention and arrange the sorting of treatment before they arrive in the emergency department. Telemedicine enables forward triage via smartphone [12], allowing physicians and patients to communicate without direct contact.

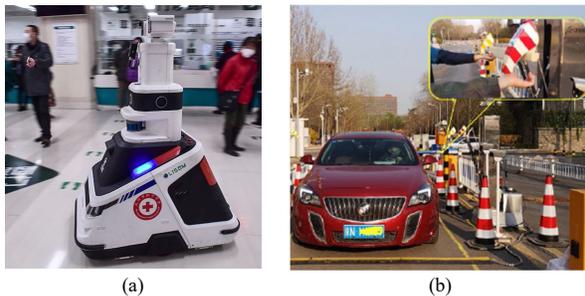

Fig. 3. Robots for screening. (a) A robot used for temperature monitoring at a hospital in Shenyang, China. (b) SHUYU robot developed by the Tsinghua University, China, allows for drive-through rapid temperature screening [13].

Incorporating thermal sensors and vision algorithms onto autonomous or remotely operated robots can increase the efficiency and coverage of screening. Robots for temperature screening were employed at the intake of many hospitals in China (Fig. 3-a). Moreover, some preexisting devices were adapted for the same purpose. Besides wheeled telerobots for indoor navigation like Diligent Robotics and Ava (iRobot Inc., USA), legged robots (Boston Dynamics, USA, Fig. 1) were adapted for telepresence and vital sign monitoring both indoor and outdoor thanks to its versatile mobility. Other robots were also deployed for drive-through rapid temperature screening, such as SHUYU robot (Tsinghua University, China, Fig. 3-b) [13], which locates human faces through computer vision and takes the temperature using a non-contact infrared thermometer.

The main challenges for improved outcomes regard the accuracy and robustness of the sensors. For instance, thermal cameras may fail in detecting the correct temperature when particular conditions are encountered, such as when sweat or a mask covers the face of the subject. Increasing the environment and context awareness of the robot would be beneficial to tackle these difficulties so that similar errors could be compensated with the aid of additional sensors and computational processing.

*C. Robots for Diagnosis*

**Bio Sampling and Image-Guided Diagnosis**

Depending on the disease, a host of samples may need to be collected, such as blood or stool samples in the case of non-airborne diseases like Ebola and cholera, or saliva, oral, or nasal swab samples in the case of airborne diseases like COVID-19. Conventional testing methods typically require interaction between a potentially infected patient and medical workers. Thus they represent occasions for the potential spreading of the virus. Additional issues may also stem from the handling of collected samples prior to, during, or after testing. Hence, robotic solutions to collecting, handling, testing, and disposing of these samples may allow for a valuable reduction in transmission of and exposure to the disease.

Telerobots with manipulation capabilities are able to achieve physical human-robot interaction, which is not feasible with conventional telemedicine solutions. As an example, the Chinese Academy of Sciences developed steerable telerobots for throat swab sampling of coronavirus tests (Fig. 4-a) [14]. During the traditional throat swab sampling, healthcare staff is in close contact with patients, which poses a high risk of cross-infection. In addition, healthcare workers' operating skills affect the accuracy and quality of swab results. To overcome those limitations, healthcare workers can teleoperate the robot with haptic feedback and visual feedback from the high-definition 3D anatomical view of binocular endoscopes. A further step was taken by a joint team from Lifeline Robotics and the University of Southern Denmark, who developed the first fully automatic throat swab robot (Fig. 1). Another work presented a portable robot (Fig. 4-b) for needle placement to draw blood or deliver fluids through an image-guided autonomous operation [15]. Multimodal image sequences

(both ultrasound and near-infrared optical imaging) were decoded by predictions from a series of deep convolutional neural networks (CNN) to guide real-time actuation of the robotic cannulation process.

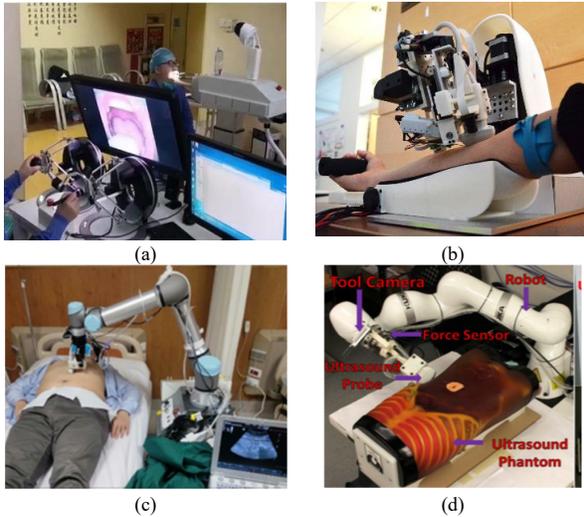

Fig. 4. Robots for sampling and diagnosis: (a) Steerable telerobot for throat swab from the Chinese Academy of Sciences[14]. (b) Portable robot exploiting deep learning for blood testing at Rutgers University, USA [15]. (c) A robotic platform for lung ultrasound at Tsinghua University, China. (d) Robotic system for remote trauma assessment at the University of Maryland, USA [16].

Another important method for diagnosis, especially during COVID-19, is the ultrasonic examination, well suited for monitoring the lung's condition, unlike the Computed Tomography (CT) scan that causes radiation and is not real-time. Tsinghua University evaluated a force-controlled ultrasound robot (Fig. 4-c) that fuses cross-modal sensory information from ultrasound and force measurements for remote diagnosis to minimize contact between healthcare staff and patients. University of Maryland developed a semi-autonomous system for hemorrhage detection using robotic ultrasound [16] and explored using the system for COVID-19 lung imaging (Fig. 4-d). A similar study was conducted to evaluate the feasibility of a remote robot-assisted ultrasound system in examining patients with COVID-19 [17]. Despite these valuable contributions, substantial technological gaps remain in dexterity, haptics, multimodal sensor integration, and autonomy, which complicate the operability of the devices. In the case of the ultrasonic examination, it turned out that training physicians to operate the robot was challenging, especially for the problem of hand-eye coordination. On the other hand, ultrasonic devices and, in general, existing medical devices are not designed for remote use by robots, so they usually need to be modified to be operated remotely.

**Lab Automation**

There is a new trend to use robots in laboratories for tasks, including sample processing, goods delivery, and conducting experiments. The primary purpose is to automate manual processes to protect personnel from infectious agents, alleviate the human workload, and achieve high throughput. The widespread COVID-19 disease has sparked a need for mass testing capacity. *Lab automation is the application most frequently requested during Ebola and COVID-19.* To address this need, the Innovative Genomics Institute (IGI) at University of California, Berkeley, established a SARS-CoV-2 testing lab in three weeks [18]. Due to the technical challenges of establishing a fully automated workflow, the IGI designed a workflow that supports two workstreams in parallel, a semi-manual approach and an automated approach in RNA extraction and liquid handling (Fig. 5), so that the testing efforts could reach the community as early as possible.

Rapid Automated BIodosimetry Tool (RABIT, Fig. 6-b) [19] developed at Columbia University is a highly automated, ultra-high throughput, biodosimetry workstation for radioactive materials handling. It can output dose estimate with no further human intervention than manual placement of the test tubes. The initial version of the RABIT system had a capacity of ~6,000 samples per day, and the goal is to reach 30,000 samples per day after parallelizing various steps. Its high throughput is partly due to pre-determined processing sequences and homogenous features of the test tubes. However, this is hardly the case in broader scenarios. For example, the mass screening of COVID-19 among a vast population led to a significant increase in the number of performed polymerase chain reaction (PCR) tests and antibody tests, which require a massive amount of heterogeneous test tubes. Despite the automation in extraction and detection, a remaining problem is the autonomous preparation of the examination plates. Osaka University [20] developed a robotic system that uses 3D vision and AI planning for autonomously arranging test tubes. Without specific instructions, the robot is able to efficiently manage the examination samples. Since the system does not require expert knowledge by human operators, it has the potential to significantly increase the throughput as well as protect and free people for more important work.

Instead of automating laboratory analysis procedures, robots can also automate the function of conducting experiments. Burger et al. used a KUKA mobile robot to automatically search for better photocatalysts for hydrogen production from water in a laboratory setting (Fig. 6-c) [21]. Thanks to its modularized approach, the robot could be used in conventional laboratories for research experiments other than photocatalysis.

Robots for laboratory automation have the potential to alleviate workload by automating manual processes and protect the personnel from being exposed to infectious agents. However, it takes significant time and resources to develop specialized robots that enable high throughput and accuracy. Thus, modular design is an opportunity to make robots adaptive to the needs of different kinds of infectious agents. Besides, all the robots need to undergo extensive benchmark and reliability tests and meet government regulations before they can be used. Per the CDC guidelines [22], the SARS-CoV-2 virus can only be cultured in laboratories with biosafety level (BSL) of 3 or higher , which significantly limits the number of facilities allowed to study the virus and thus hinders the development of new treatments or vaccines. All these obstacles affect how soon robotic solutions can be available to the public.

|  | Extract RNA | RT-qPCR | Analyze data | Interpret data |
|---|---|---|---|---|
| Thermo Fisher EUA protocol | Collect patient samples followed by manual or automated RNA extraction | Perform RT-qPCR | Analyze data using commercial software | Interpret result based on Thermo Fisher's EUA |
| Phase 1 Semi-automated | **Manual** RNA extraction from previously arrayed patient samples using an automated liquid handling robot | Manual RT-qPCR plate setup, perform RT-qPCR | Analyze data using **custom software** | Interpret result based on Thermo Fisher's EUA **and criteria established in IGI's validation assays** |
| Phase 2 Automated | **Automated** RNA extraction from previously arrayed patient samples using an automated liquid handling robot | **Automated** RT-qPCR plate setup by a liquid handling robot, perform RT-qPCR | Analyse data using custom software | Interpret result based on Thermo Fisher's EUA and criteria established in IGI's validation assays |

Fig. 5. The workflow of manual and automated protocols proposed by the Innovative Genomics Institute (IGI) at the University of California, Berkeley [18] (blue background). Their implementation of the workflow is built upon Thermo Fisher's authorized Emergency Use Authorization (EUA) protocol [23] (yellow background). Phase 1 of the IGI's workflow requires the manual implementation of the Thermo Fisher kit, while Phase 2 is automated. Bolded words indicate elements changed from the implementation above. Icons credit: Flaticon.com.

*D. Robots for Treatment*

One in every six patients with COVID-19 experienced severe conditions involving bilateral pneumonia and acute respiratory distress syndrome [24]. Therefore, endotracheal intubation was one of the most required treatments to allow for mechanical ventilation. Intubation is a complicated procedure with high complication rates, which strongly relies on the manual dexterity of experienced physicians [25]. It is performed by placing a tracheal tube into the trachea of the patient while lifting the jaw with a laryngoscope. This procedure implies direct contact with contagious airways, and in the situation of infectious disease like that of COVID-19, it exposes the operator to a high risk of infection. Robots, such as protective devices, can provide valuable help to ensure the safety of doctors and patients during operation, especially in emergency situations. As shown in Fig. 7, intubation teams involved in airway management procedures are composed of several members with defined roles. Thus a robot can potentially assist or replace some personnel, relieving the annexed workflow and reducing the probability of disease transmission.

Researchers at John Hopkins University designed a Cartesian robot with an integrated camera that enables remote control of a ventilator [26]. Airway management in infectious conditions is a challenge that may greatly benefit from robotic assistance to make the intubation safer and more efficient.

Nonetheless, unintended issues and undesired consequences may arise from their deployment, like reducing the ergonomics and maneuverability of operators or the introduction of additional contamination sources. Therefore, it is critically important that the use of any additional tool is adequately considered and handled [27] [28]. Researchers explored general-purpose surgical robotic platforms, such as the Da Vinci surgical system [29], and purposely designed devices to address specific issues. Ad-hoc solutions span across the entire spectrum of robotics, from fully manual teleoperation to assisted sensing and actuation, up to the eventual deployment of fully autonomous systems [30]. One interesting result is the development of intubation systems remotely controlled via a joystick by the user. This strategy was embodied both in a fixed platform (Fig. 8-a) [29] and a portable device, Remote Robot-Assisted Intubation System [31] for hospital and pre-hospital treatment, respectively. Yet, the main challenges of robot-assisted intubation are related to the lack of tactile feedback. Researchers from Columbia University developed InTouch, an advanced laryngoscope with a tactile sensing blade [32]. Even though it demonstrated to decrease the complication rate and the time required for correct intubation, it does not include any automation feature and still relies on visual recognition of the airway and manual steering by its user. A step further in this direction is taken with REALITI, a Robotic Endoscope Automated via Laryngeal Imaging for Tracheal Intubation (Fig. 8-b), developed at ETH Zurich [33]. It handles the task of guiding the tracheal tube into its correct position by performing the automated detection of anatomical landmarks within the throat and the automated steering of endoscopes toward the recognized features. While this system has been successfully tested on manikins, a long way is still required before it can be effectively deployed on human subjects.

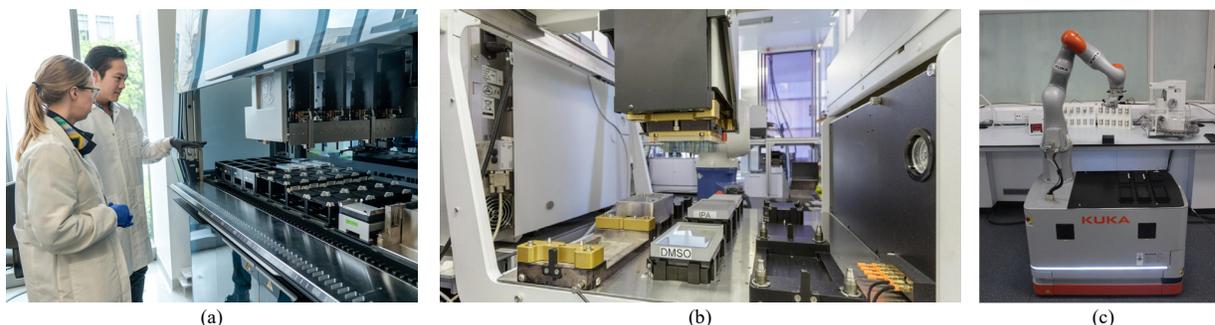

(a)           (b)           (c)

Fig. 6. Robots for laboratory automation. (a) Automated liquid-handling robot (Hamilton STARlet) at University of California Berkeley will be used to analyze swabs from patients to diagnose COVID-19 [18]. (b) A rapid automated biodosimetry tool (RABIT) from Columbia University [19]. (c) A mobile robotic chemist for conducting experiments [21].

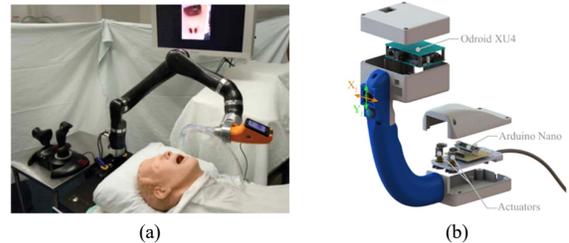

(a) (b)

Fig. 8. Robots for intervention and treatment: (a) Kepler Intubation System (McGill University, Canada) [29]. (b) Robotic Endoscope Automated via Laryngeal Imaging for Tracheal Intubation (REALITI, ETH Zurich, Switzerland) [33].

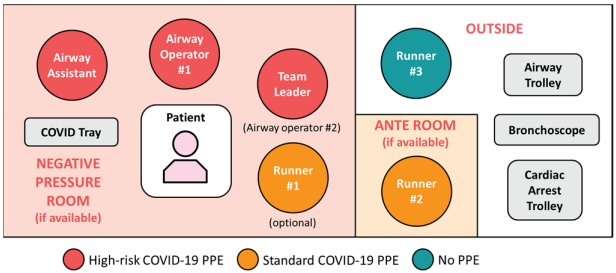

Fig. 7. According to the Safe Airway Society principles of airway management and tracheal intubation, intubation teams are composed of multiple operators with specific roles. Hence, the adoption of robotic solutions may substantially relieve the workflow by replacing/assisting some personnel. [34].

A key challenge is the necessity of a robust method for accurately identifying the anatomical features in a large and multifaceted population, whereby variations in airways' anatomy, local pathologies, and other particular conditions may hinder regular tracheal intubation. To this extent, significant improvements might derive from the constantly progressing tools of computer vision and artificial intelligence. Other opportunities come from the enhancement of sensing and actuation technologies. Besides visual feedback, force sensing is crucial to enhance the reliability of robotic intubation. Its success rate could significantly benefit from distributed, and accurate force/pressure sensors made possible by advancements in stretchable electronics. From the actuation perspective, a promising approach could derive from the conquers of soft robotics. A vine-inspired robot, for instance, may represent a feasible way to create a conduit to the lungs via a failsafe and easy-to-use device, with the ability to grow into multiple branches to deal with different morphologies [35].

*E. Robots for Homecare*

The COVID-19 pandemic caused an overwhelming load of the health facilities requiring intensive employment of the hospital workforce. Doctors and nurses strive to work extra shifts to save lives [36]. Additionally, direct contact with infectious patients puts them in the critical condition of being regularly exposed to the risk of contracting the disease. Therefore, robots capable of performing typical assistance tasks would enormously benefit the daily medical care of patients during similar circumstances.

Not only in hospitals, nursing assistance is also a fundamental service in nursing homes and domestic spaces. It covers all the workflow of care, from the acceptance of the patient to the hospital until the recovery and aid service at home. To date, there are no robots that can effectively carry out the versatile activities of nurses. The great challenge remains the ability to deal with a plethora of tasks and a myriad of different subjects, requiring both physiological empathy and physical interaction. The monitoring of patients, delivery of meals and medication, assistance with patient ambulation, and manipulation of medical equipment are only a few examples highlighting the incredible amount of required versatility.

Most nursing robots are only able to perform very basic functions, e.g., telepresence and meal delivery. An example is the Sanbot Elf robot (Qihan Technology, China), which became famous as Tommy while treating COVID-19 patients in an Italian hospital (Fig. 9-a). Another class of nurse robots embraces robots for social assistance [37]. Indeed, psychological support is also critical during pandemic emergencies, when mental health is exacerbated by severe restrictions, such as quarantine and social distancing. More generally, five primary functions have been identified for general-purpose nursing robots: communication, mobility, measurement of clinical data, general manipulation, and tool use [38]. Among these, physical interaction constitutes the key challenge, and in particular, manipulation is the bottleneck towards effective deployment of nursing robots due to the necessity of a broad range of dexterity and strength, involving dealing with both gross, powerful actions (e.g., patient assistance during lifting and walking) and delicate, precise manipulation (e.g., intravenous fluid management). Advances in soft robotics enable substantial progress in this direction, with the development of numerous adaptive and versatile grippers. In addition to the end-effector solution, high torque density actuators demonstrated high compliance and high bandwidth in legged and wearable robots [39]. They can be incorporated into a robot arm design to enhance safety and performance during the interaction.

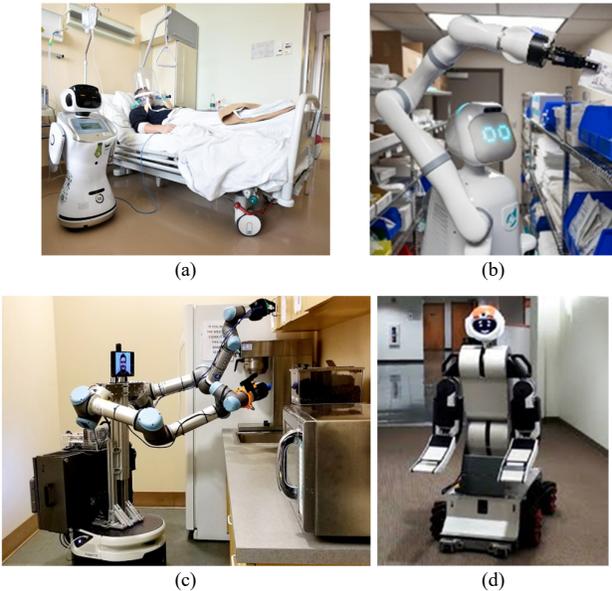

(a) (b) (c) (d)

Fig. 9. Robots for nursing assistance: (a) Sanbot Elf robot (Qihan Technology, China) worked as a nurse nicknamed Tommy in an Italian hospital. (b) Moxi (Diligent Robotics Inc., USA). (c) Tele-robotic Intelligent Nursing Assistant (TRINA, University of Illinois at Urbana-Champaign & Duke University, USA) [40]. (d) Robotic Nursing Assistant (RoNA, Hstar Technologies, USA) [41].

Another challenge is related to the use of tools since medical devices are not designed for being used by robots. An adopted solution is the use of special replaceable connectors for various grippers to reconfigure the end effector [42]. To tackle these challenges, the baseline approach relies on the employment of traditional anthropomorphic robotic arms installed on omnidirectional mobile platforms. Additional features include different levels of autonomy, autonomous 3D mapping and navigation, telepresence with two-way audio for telerobotic-human communication, patient surveillance, and assistance for patient mobility (e.g., during lifting and walking).

Some of the most advanced embodiments of these technologies include both commercial products, like Moxi (Diligent Robotics, USA, Fig. 9-b), and research prototypes, such as the Tele-robotic Intelligent Nursing Assistant (University of Illinois at Urbana-Champaign, USA, Fig. 9-c) [40], Robotic Nursing Assistant (Hstar Technologies, USA, Fig. 9-d) [41], and a telerobotic system based on YuMi (ABB, Switzerland) [42]. The latter was developed for remote care operation in the isolation ward and was tested in the Hospital of Zhejiang, China. The robot is composed of two subsystems, one for telepresence and one for teleoperation, and can assist or even replace the medical staff with taking care of patients in tasks that include daily checkups, delivery of medicine, food, or other essentials, operation of medical instruments, disinfection of frequently touched surfaces, and auscultation while wearing PPE.

## IV. DISCUSSION AND CONCLUSION

Robotic systems are currently being used to perform more-and-more sophisticated tasks throughout our society. It is not surprising that they are playing an important role in responding to the challenges posed by the current COVID-19 pandemic. In this survey, we have focused primarily on robotic applications in healthcare, including those directly related to the care of COVID-19 patients in hospitals and those allowing provision of ordinary care at homes (e.g., nursing robots). Specifically, prevention, screening, diagnosis & treatment in hospital settings, and post-recovery home care are discussed. Other application areas, such as public safety, supply chain logistics, and transportation, are also important. Indeed, the adaptability of robot systems and technology means that there has been considerable cross-talk between areas. Many healthcare robots discussed in this survey are essentially adaptations of robotic systems developed for non-healthcare uses.

As this paper is being written, the pandemic continues, and robotic systems may be expected to play an increasing role in dealing with the challenges presented. However, it is not too soon to consider some of the lessons drawn from the experience.

The first lesson is the crucial role of *communication* between the user communities that are most immediately affected by the disease and the engineers and robotics researchers who are developing systems to address emerging needs. Unless there is a good understanding of the unmet needs and the constraints imposed by the environment into which a robot is to be introduced, it is not likely that it will be useful. Similarly, it is important to understand the differing needs and expectations of all the people (technicians, healthcare workers, patients, family members) with whom the system is likely to interact.

A second lesson is the importance of *capability* and *adaptability* in robotic systems. As mentioned above, many of the systems discussed in the paper adapt robotic capabilities developed for other uses to meet emergent needs. This adaptability is almost an inherent aspect of robotic systems, and the trend will continue, both due to research programs such as the US National Robotics Initiative (NRI) and increasing commercial deployment of robots. It is crucial for those involved in developing these systems to remain sensitive to the importance of future flexibility while also concentrating on the demands of safety, simplicity, and robustness in meeting a current requirement.

A third lesson concerns *deployability* in sufficient numbers to make a major difference in a crisis. Although the systems we discussed meet real needs, only a relatively small number have actually been installed. As more robots are installed broadly in our economy, there is at least the potential to exploit their inherent adaptability to be put to work in healthcare applications. However, advanced planning and preparation for such a mobilization seem important.

A final lesson concerns the need for better preparation for the *infectious disease-specific constraints* associated with operating in a pandemic environment. These include such matters as cleaning and disinfection protocols and materials choices.

The adoption of technology needs to be expedient but safe and *responsible* for facing disasters like a pandemic crisis.

Speedy and mindful regulations that properly weigh the benefits and risks are necessary to guarantee the safety and the effectiveness of robots and prevent biases and privacy issues. Joint efforts by roboticists, government, industry, and citizen stakeholders may indeed facilitate the development and deployment of useful and validated robotic solutions for the benefit of the community. If appropriate strategies are implemented to ensure adaptable and reliable systems that can be quickly replicated and distributed on demand, robots could play a much more significant role in future crises.